\documentclass{article}

\PassOptionsToPackage{numbers, compress}{natbib}
\usepackage{graphicx}
\usepackage[preprint]{neurips_2025}

\usepackage{enumitem}
\usepackage[utf8]{inputenc} % allow utf-8 input
\usepackage[T1]{fontenc}    % use 8-bit T1 fonts
\usepackage{hyperref}       % hyperlinks
\usepackage{url}            % simple URL typesetting
\usepackage{booktabs}       % professional-quality tables
\usepackage{amsfonts}  
\usepackage{wrapfig}
\usepackage{nicefrac}       % compact symbols for 1/2, etc.
\usepackage{microtype}      % microtypography
\usepackage{xcolor}         % colors
\usepackage{amsmath} 
\usepackage{algorithmic}
\usepackage{algorithm}
\usepackage{multirow}
\usepackage{enumitem}
\newcommand{\ours}{ALinear}
\newcommand{\ourslong}{Adaptive Linear}

\definecolor{darkblue}{RGB}{0, 0, 140}
\definecolor{lightblue}{RGB}{0, 128, 255}

\title{Does Scaling Law Apply in Time Series Forecasting?}

\author{%
    Zeyan Li\\
    Jinan University\\
    \texttt{zeyan0823@gmail.com} \\
    \And
    Libing Chen \\
    Jinan University \\
    \texttt{clb2004prc@gmail.com} \\
    \And
    Yin Tang* \\
    Jinan University \\
    \texttt{ytang@jnu.edu.cn} \\
}

\begin{document}

\maketitle

\begin{abstract}
Rapid expansion of model size has emerged as a key challenge in time series forecasting. From early Transformer with tens of megabytes to recent architectures like TimesNet with thousands of megabytes, performance gains have often come at the cost of exponentially increasing parameter counts. But is this scaling truly necessary? To question the applicability of the scaling law in time series forecasting, we propose Alinear, an ultra-lightweight forecasting model that achieves competitive performance using only k-level parameters. We introduce a horizon-aware adaptive decomposition mechanism that dynamically rebalances component emphasis across different forecast lengths, alongside a progressive frequency attenuation strategy that achieves stable prediction in various forecasting horizons without incurring the computational overhead of attention mechanisms. Extensive experiments on seven benchmark datasets demonstrate that Alinear consistently outperforms large-scale models while using less than 1\% of their parameters, maintaining strong accuracy across both short and ultra-long forecasting horizons. Moreover, to more fairly evaluate model efficiency, we propose a new parameter-aware evaluation metric that highlights the superiority of ALinear under constrained model budgets. Our analysis reveals that the relative importance of trend and seasonal components varies depending on data characteristics rather than following a fixed pattern, validating the necessity of our adaptive design. This work challenges the prevailing belief that larger models are inherently better and suggests a paradigm shift toward more efficient time series modeling.
\end{abstract}

\section{Introduction}
\label{sec:introduction}

With increasingly complex deep learning architectures outperforming classical statistical methods, time series forecasting has undergone a paradigm shift. From recurrent neural networks to even sophisticated transformer-based models \cite{vaswani2017attention, zhou2021informer, wu2021autoformer}, this trend, accompanied by an exponential growth in parameters, raises a fundamental question that motivates our research: \textit{Is it necessary for a large number of parameters to yield better performance in time series forecasting?}

The time series forecasting landscape has undergone a profound transformation over the past decade, characterized by increasing architectural complexity and parameter counts. 
The neural network era began with RNNs \cite{elman1990finding} and LSTMs \cite{graves2012long}, which introduced non-linear modeling capabilities but remained relatively parameter-efficient. 
Exemplified by Transformer~\cite{vaswani2017attention}, a significant paradigm occurred with the adaptation of attention mechanisms to temporal data. This shift catalyzed a wave of sophisticated temporal architectures: ProbSparse self-attention mechanism~\cite{zhou2021informer}, decomposition techniques with attention~\cite{wu2021autoformer}, frequency-domain transformations~\cite{zhou2022fedformer}. Models like TimesNet~\cite{wu2022timesnet} have exceeded above 1,000M parameters in pursuing marginal performance gains disproportionate to the increased model size. 
Parallel to this architectural inflation, a counter-current of research has emerged questioning the necessity of such complexity. DLinear~\cite{zeng2023are} demonstrated competitive performance with dramatically reduced parameter counts by employing simple decomposition-based linear projections. 
These findings collectively indicate a fundamental misalignment between architectural complexity and the intrinsic nature of time series forecasting problems.

\begin{figure}[htbp] 
\centering
\includegraphics[width=1\linewidth]{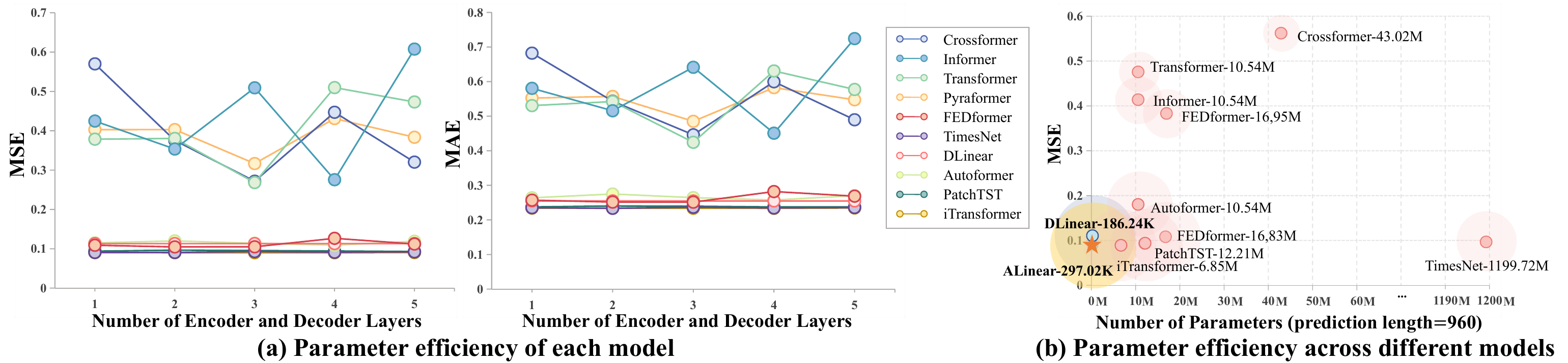}
\vspace{-1em}
\caption{Comparison of model's parameter efficiency on ETTm1 dataset when prediction length = 960.  Figure \ref{fig:introduction} (a) shows the metrics below different number of layers of Encoder/Decoder of the state-of-the-art (sota) time series forecasting model. Figure \ref{fig:introduction} (b) illustrates the comparison of sota models in terms of performance and parameter quantity.}
\label{fig:introduction}
\vspace{-1em}
\end{figure}

The success of scaling law in natural language processing and computer vision \cite{kaplan2020scaling,zhai2022scaling} has led to the assumption that it would also apply to time series forecasting. However, this is not the case. For time series forecasting tasks, we have conducted numerous experiments on the encoder and decoder network layers of each model, as shown in Figure \ref{fig:introduction} (a). The results indicate that such operations have not significantly enhanced model performance. This suggests that the solution to improving performance may not lie in further parameter expansion but in rethinking how models adapt to the inherent properties of time series.

Another fundamental challenge in time series forecasting, which we term as the "Forecasting Horizon Dilemma" \cite{hyndman2018forecasting}, manifests as a dramatic deterioration in predictive accuracy—often exceeding large degradation when moving from short-term to ultra-long-term forecasting. The dilemma stems from an oversight in current approaches: the failure to recognize that the nature of prediction changes fundamentally as the horizon extends. Time series theory establishes that different temporal components exhibit varying predictability across time scales \cite{hamilton2020time,percival2000wavelet}. Trend components maintain predictability over longer horizons, while seasonal patterns become increasingly stochastic \cite{hyndman2018forecasting}. High-frequency components that are informative for short-term forecasting become noise when predicting further into the future \cite{li2023revisiting,shi2024time}.

Beyond theoretical debates, the trend of parameter expansion has posed significant practical challenges: resource-heavy training processes restrict accessibility, inference latency constraints prevent many state-of-the-art models from being deployed in real-world applications, and environmental concerns surrounding the training of large models are mounting \cite{benidis2022deep,strubell2020energy,schwartz2020green}. 

Motivated by these insights, we present \ourslong~(\ours), a model that fundamentally challenges both the scaling law of parameters and the static-horizon forecasting approach with much fewer parameters than popular modern models through adaptive decomposition, trend-seasonal balancing, and progressive frequency decay mechanisms. Specifically, the contributions extend current scientific understanding as follows:
\begin{itemize}[leftmargin=*]
    % \vspace{-2.5em}
    % \setlength{\itemsep}{0pt}
    % \setlength{\parsep}{0pt}
    % \setlength{\parskip}{0pt}
    \item We challenge the parameter-scaling stereotype in time series forecasting by demonstrating experiments of various amount of parameters across state-of-the-art models, and by proposing a simple yet powerful model with much fewer parameters. The results suggest that the complexity of effective forecasting may be significantly lower than previously assumed. 
    
    \item We propose ALinear, which can achieve sota performance across multiple benchmarks with just k-level parameters — lower than 1\%  parameters of comparable models (Figure \ref{fig:introduction} (b)). Its performance, at the same time, keeps extremely stable under different hyperparameter combination and is suitable for deployment to real-world production applications.

    \item To more fairly evaluate model efficiency, we propose a new parameter-aware evaluation Parameter-Normalized Performance (PNP) metric which can show the parameter efficiency.
\end{itemize}

The remainder of this paper is organized as follows: Section \ref{sec:related} contextualizes our work within broader literature landscape, Section \ref{sec:approach} details the \ours\ architecture, Section \ref{sec:experiments} provides empirical validation across diverse benchmarks, and Section \ref{sec:conclusion} discusses implications and future research directions.

\section{Related work}
\label{sec:related}

\subsection{Decomposition and horizon-aware methods}

Time series decomposition separates data into distinct components (trend, seasonality, residual), enabling specialized processing techniques\cite{li2025faith}. Classical methods like STL \cite{rb1990stl} established systematic extraction approaches, while models like Prophet \cite{taylor2018forecasting} extended these concepts with Bayesian methods for robust decomposition. Neural decomposition emerged with N-BEATS \cite{oreshkin2019n}, which learns trend and seasonal components through specialized blocks, while Autoformer \cite{wu2021autoformer} integrated decomposition with Transformer architectures. Horizon-aware approaches explicitly incorporate prediction length into modeling, recognizing that different forecast distances exhibit different statistical properties \cite{zhou2022film}. Traditional methods \cite{makridakis2018statistical} inherently account for prediction distance through recursive structures, while direct forecasting methods \cite{marcellino2006comparison} train separate models for each horizon. In deep learning, recurrent models naturally adapt to different horizons but suffer from error accumulation. Advanced models like Informer \cite{zhou2021informer} and PatchTST \cite{nie2022time} implement specialized mechanisms for long-sequence forecasting and horizon-specific processing. Despite these innovations, current decomposition methods typically apply fixed strategies regardless of prediction horizon, contradicting established understanding that different forecast horizons require different emphasis on trend and seasonal components \cite{hyndman2018forecasting}. 
%\textcolor{red}{In contrast to these approaches, \ours introduces a horizon-adaptive decomposition framework that dynamically adjusts decomposition parameters based on prediction length. This allows our model to optimize the extraction and processing of trend and seasonal components specifically for each forecast horizon, addressing a critical gap in existing decomposition methods while maintaining extreme parameter efficiency.}

\subsection{Efficient architectures and linear models}

Parameter-efficient forecasting models achieve high accuracy with minimal computational complexity, challenging assumptions that increased model size necessarily improves performance. TCNs \cite{bai2018empirical} used dilated convolutions for wide receptive fields with fewer parameters, while Transformer variants \cite{kitaev2020reformer,beltagy2020longformer} introduced efficiency mechanisms through hashing and sparse attention. A notable paradigm shift occurred with DLinear \cite{zeng2023are}, demonstrating that simple linear layers applied to decomposed time series could outperform complex Transformers with orders of magnitude fewer parameters. This sparked renewed interest in architectural simplicity, with models like TiDE \cite{das2023long} achieving competitive performance through lightweight transformations. These findings suggest that time series data may be effectively captured through well-designed linear models coupled with appropriate decomposition strategies \cite{li2023revisiting}. However, many linear models lack mechanisms to adjust behavior based on prediction distance, potentially oversimplifying near-term patterns while overfitting to noise in longer horizons. 
%\textcolor{red}{\ours builds upon these efficiency insights while addressing their limitations by combining extreme parameter efficiency (297.02K parameters) with horizon-adaptive mechanisms. Our approach demonstrates that not only can simple architectures match or exceed the performance of complex models, but they can also incorporate sophisticated adaptation strategies without sacrificing simplicity or efficiency.}

\subsection{Emerging paradigms: frequency domain and pre-trained models}

Frequency domain methods offer distinct advantages in capturing periodic patterns and long-range dependencies. FreTS \cite{yi2023frequency} demonstrated that frequency-domain MLPs can outperform time-domain models through natural alignment with periodic patterns, while hybrid spectral-temporal approaches leverage complementary advantages from both domains\cite{zhou2022fedformer}. Recent innovations like TimesNet \cite{wu2022timesnet} transform 1D time series into 2D representations across multiple periods, while MICN \cite{wang2023micn} employs multi-scale architectures to model different frequency components. Additionally, frequency domain methods typically apply uniform processing across all frequency components regardless of prediction horizon. 

Following successes in NLP and computer vision, pre-trained models have emerged for time series forecasting. Approaches like TimeLLM \cite{jin2023time} adapt large language models through specialized prompting, while TimeGPT-1 \cite{garza2023timegpt} represents one of the first foundation models trained specifically for cross-domain forecasting. Recent innovations include TimeFM \cite{das2024decoder} establishing blueprints for domain adaptation and Chronos \cite{ansari2024chronos} leveraging contrastive pre-training. These approaches offer promising directions for knowledge transfer across domains, addressing data scarcity issues in forecasting applications. While these emerging paradigms offer valuable insights, they often introduce significant computational complexity, parameter inefficiency, and worse interpretability. 

It is worth noting that, many state-of-the-art forecasting models rely heavily on careful hyperparameter tuning to maintain accuracy across different settings. Works such as Autoformer \cite{wu2021autoformer}, Crossformer \cite{zhang2023crossformer}, and PatchTST \cite{nie2022time} often involve complex architectural choices. Additionally, recent studies in large-scale time series models such as TimeGPT \cite{garza2023timegpt} and Lag-Llama\cite{rasul2023lag} suggest that even minor changes to the prompt text can lead to substantial degradation in performance due to the brittleness of foundation model architectures in temporal settings. 

\section{Adaptive Linear Network}
\label{sec:approach}

\begin{figure*}[t]
\centering
\includegraphics[width=\linewidth]{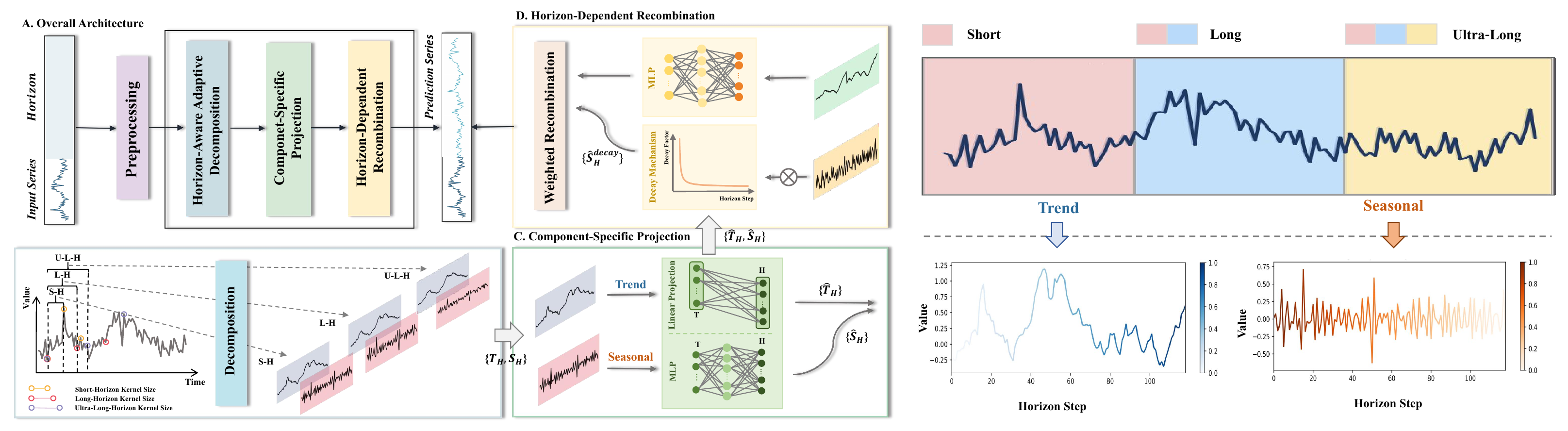}
\vspace{-1em}
\caption{\ours\ model. The model separates the input time series into trend and seasonal components through a decomposer with adaptive kernel size, then separately through component-specific projections, and combines the predictions through adaptive trend-seasonal balancing. The right side of the figure reflects our design concept. As the predicted length varies, the importance of periodicity and trend (the depth of the colors in the figure) is constantly changing.}
\label{fig:model_architecture}
\vspace{-1em}
\end{figure*}

This section presents our proposed \ourslong~(\ours). Formally, given a univariate time series \(\mathbf{X} = \{x_1, x_2, \ldots, x_{T}\} \in \mathbb{R}^{T}\), where \({T}\) represents the input length, the prediction task aims to forecast the next \(H\) time steps, obtaining \(\mathbf{Y} = \{x_{{T}+1}, x_{{T}+2}, \ldots, x_{T+H}\} \in \mathbb{R}^{H}\), where \(H\) is the forecast horizon. Figure \ref{fig:model_architecture} illustrates \ours's architecture, whose components we describe below.

\subsection{Horizon-Aware Adaptive Decomposition}
\label{subsec:decomposition}

Different forecast horizons require different treatment of underlying components \cite{hyndman2018forecasting}. Short-term forecasts benefit from capturing fine-grained seasonal patterns, while long-term forecasts should prioritize stable trend components and attenuate noise. Despite this established principle, most existing decomposition approaches apply fixed strategies regardless of the prediction task. We construct a \emph{Horizon-Adaptive Decomposition} framework that dynamically adjusts based on the forecast horizon \(H\), formalizing this through a horizon-dependent decomposition function $\mathbf{D}_H$:
\begin{equation}
\mathbf{D}_{H}(\mathbf{X}) \rightarrow \mathbf{T}_{H}, \mathbf{S}_{H},
\end{equation}
where \(\mathbf{T}_{H}, \mathbf{S}_{H} \in \mathbb{R}^T\) represent trend and seasonal components extracted with parameters specifically optimized for horizon \({H}\). We implement this using a horizon-parameterized moving average with adaptive kernel size:
\begin{equation}
\mathbf{T}_{H} = \text{MA}_{\alpha({H})}(\mathbf{X}), \mathbf{S}_{H} = \mathbf{X} - \mathbf{T}_{H},
\end{equation}
where \(\text{MA}_{\alpha({H})}\) is a moving average with window size determined by \(\alpha({H})\). The critical innovation lies in our window size function:
\begin{equation}
\alpha({H}) = \min\left(\max\left(k_1 + k_2 \cdot {H}, w_{\text{min}}\right), w_{\text{max}}\right),
\end{equation}
with learnable parameters \(k_1, k_2\) and hyperparameters \(w_{\text{min}}, w_{\text{max}}\) that constrain the window size within practical bounds. This formulation creates a continuous, differentiable relationship between forecast horizon and decomposition strategy, allowing the model to learn the optimal decomposition for each prediction length through gradient-based optimization.

\subsection{Component-Specific Projections}
\label{subsec:projections}

After adaptive decomposition, \ours\ employs specialized linear projections for each component, following the principle that trend and seasonal patterns require different modeling approaches. This design choice is supported by statistical theory that suggests different time series components exhibit distinct mathematical properties and should be modeled separately \cite{rb1990stl, taylor2018forecasting}.

The \emph{Component-Specific Projections} are defined as:
\begin{equation}
\hat{\mathbf{T}}_{H} = \mathbf{W}_{T} \mathbf{T}_{H} + \mathbf{b}_{T}, \hat{\mathbf{S}}_{H} = \mathbf{W}_{S} \mathbf{S}_{H} + \mathbf{b}_{S},
\end{equation}
where \(\mathbf{W}_{T}, \mathbf{W}_{S} \in \mathbb{R}^{H \times T}\) are learnable weight matrices and \(\mathbf{b}_{T}, \mathbf{b}_{S} \in \mathbb{R}^H\) are bias vectors for trend and seasonal components. Unlike complex architectures that apply identical transformations to different components, our approach allows each projection to specialize in capturing the unique characteristics of its respective component. The trend projection \(\mathbf{W}_{T}\) can focus on long-term dependencies and smooth transitions, while the seasonal projection \(\mathbf{W}_{S}\) can capture recurring patterns and cyclical behaviors. This specialization is achieved without increasing model complexity, as the linear projections maintain parameter efficiency while providing sufficient expressivity for component-specific modeling.

\subsection{Horizon-Dependent Recombination}
\label{subsec:recombination}

The final stage of \ours\ implements a theoretically-motivated, \emph{Horizon-Dependent Recombination} mechanism that addresses a fundamental challenge in time series forecasting: the importance of different ingredients is constantly changing. The horizon-adaptive decay to the predicted seasonal output is calculated as: 
\begin{equation}
\hat{\mathbf{S}}_{H}^{\text{decay}}(t) = \hat{\mathbf{S}}_{H}(t) \cdot \exp(-\lambda \cdot t),
\end{equation}
where \(\hat{\mathbf{S}}_{H}(t) \in \mathbb{R}^{H}\) is the seasonal prediction at horizon step \(t \in \{1, 2, \ldots, {H}\}\) and \(\lambda\) is a decay rate determined by the prediction length \({H}\). Specifically, we define \(\lambda = \delta/{H}\), where \(\delta\) is a hyperparameter, depending on whether the task is short-term, long-term, or ultra-long-term forecasting, respectively. This leads to a smooth, exponential attenuation of seasonal components over time, allowing the model to discount unreliable high-frequency signals in distant predictions while retaining near-future periodic patterns. Finally, our adaptive recombination is formulated as:
\begin{equation}
\hat{\mathbf{Y}} = \beta_{T}({H}) \cdot \hat{\mathbf{T}}_{H} + \beta_{S}({H}) \cdot \hat{\mathbf{S}}_{H}^{\text{decay}},
\end{equation}
where \(\beta_{T}({H}) = \sigma(v_1 + v_2 \cdot {H})\) and \(\beta_{S}({H}) = 1 - \beta_{T}({H})\), with \(\sigma\) being the sigmoid function and \(v_1, v_2\) learnable parameters that control the horizon-dependent balancing. This formulation creates a continuous and differentiable relationship between the forecast horizon and the importance of the components. As \({H}\) increases, \(\beta_{T}({H})\) approaches 1, allocating more weight to trend components that maintain predictive power over longer horizons. As ${H}$ decreases, \(\beta_{S}({H})\) increases, emphasizing seasonal components which provide valuable short-term signals. This progressive frequency decay mechanism effectively implements a learnable spectral filter that attenuates high-frequency components as the prediction horizon extends, addressing the "Forecasting Horizon Dilemma" we identified in Section \ref{sec:introduction}.

\subsection{Training procedure and optimization}
\label{subsec:training}

\begin{figure*}[t]
\centering
\includegraphics[width=0.99\linewidth]{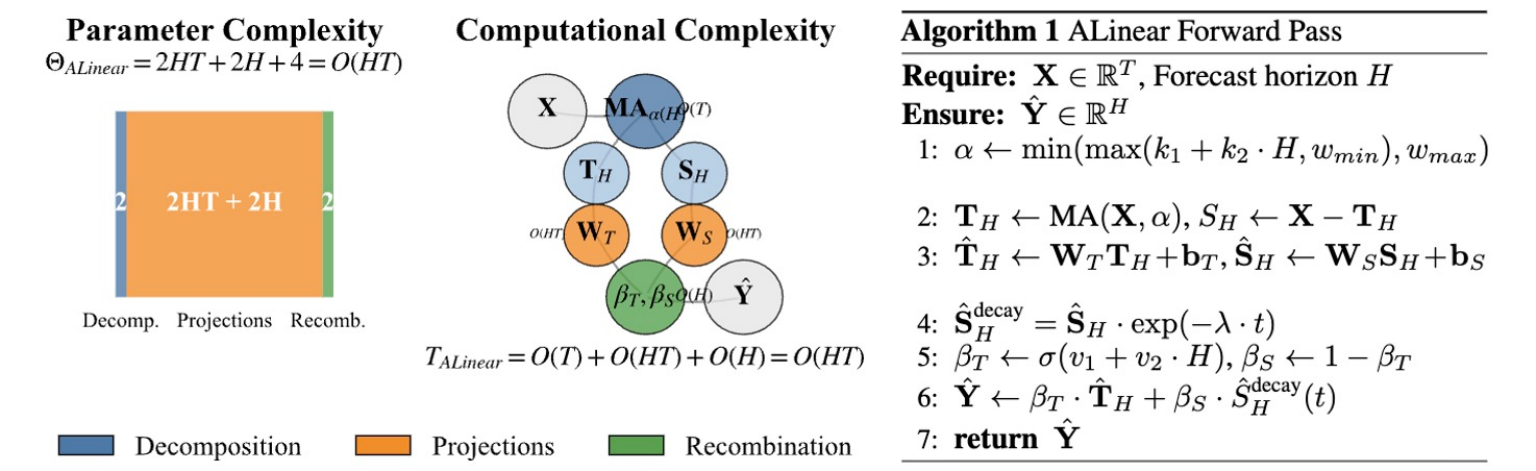}
\vspace{-0.5em}
\caption{Efficiency analysis and forward pass algorithm of \ours}
\vspace{-1em}
\label{fig:algorithm}
\end{figure*}

\ours~employs an end-to-end training approach that simultaneously optimizes all horizon-adaptive components through gradient descent. The model is trained to minimize the mean squared error between predictions \(\hat{\mathbf{Y}}\) and ground truth \(\mathbf{Y}\):
\begin{equation}
\mathcal{L} = \frac{1}{{H}} \sum_{i=1}^{{H}} (y_i - \hat{y}_i)^2.
\end{equation}
This unified optimization enables the model to learn complex interdependencies between the decomposition strategy, component-specific projections, and recombination weights. The learnable parameters include adaptive decomposition parameters \(k_1, k_2\) that control the horizon-dependent window sizes, projection parameters \(\mathbf{W}_{T}, \mathbf{W}_{S}, \mathbf{b}_{T}, \mathbf{b}_{S}\) that capture component-specific patterns, and recombination parameters \(v_1, v_2\) that balance component contributions based on forecast horizon. Figure \ref{fig:algorithm} presents \ours's forward pass, highlighting the elegant simplicity of the approach.

\subsection{Efficiency analysis}
\label{subsec:complexity}

A key contribution of \ours~is achieving state-of-the-art forecasting performance while maintaining exceptional computational and parameter efficiency. The total parameter count of \ours\ is precisely quantified as:
\begin{equation}
\Theta_{\text{ALinear}} = \underbrace{2}_{\text{decomp.}} + \underbrace{2{HT} + 2{H}}_{\text{projections}} + \underbrace{2}_{\text{recomb.}} = 2{HT} + 2{H} + 4.
\end{equation}

For typical forecasting scenarios where \({T} \ll {H}\), this simplifies to \(\mathcal O(\text{HT})\). The time complexity for \ours's forward pass is precisely:
\begin{equation}
\begin{aligned}
T_{\text{ALinear}} &= \underbrace{\mathcal O({T})}_{\text{decomp.}} + \underbrace{\mathcal O({HT})}_{\text{projections}} + \underbrace{\mathcal O({H})}_{\text{recomb.}} = \mathcal O({HT}).
\end{aligned}
\end{equation}

The space complexity is \(\mathcal O({T + H})\) for storing inputs and outputs, plus \(\mathcal O({HT})\) for parameters, resulting in total space complexity of \(\mathcal O({HT})\).  This efficiency makes \ours\ particularly suitable for resource-constrained environments and real-time applications.

\section{Experiments}
\label{sec:experiments}

In this section, we evaluate the performance of the \ours.
Extensive experiments carried out on seven real-world datasets are designed to explore and answer the following research questions ({\bf RQs}):
\begin{description}[labelwidth=2em, labelsep=1em, leftmargin=3em]
    \vspace{-1em}
    \setlength{\itemsep}{0pt}
    \setlength{\parsep}{0pt}
    \setlength{\parskip}{0pt}
    \item[\textbf{RQ1}] What is the overall performance of \ours~in long-term time series forecasting?
    \item[\textbf{RQ2}] How different are our parameter advantages compared with the existing sota models?
    \item[\textbf{RQ3}] How do the internal parameters of the \ours~change as the prediction length increases?
    \item[\textbf{RQ4}] What impact do the various components have on time series forecasting and whether or not the hyper-parameter of \ours~is sensitive?
\end{description}

\textbf{Experimental setup.}
We carefully selected a diverse suite of benchmark datasets  (\textbf{ETT datasets} (ETTh1, ETTh2, ETTm1, ETTm2), \textbf{Exchange Rate}, \textbf{Traffic}, \textbf{Weather}) to validate our hypotheses across different domains, temporal granularities and pattern complexities. To comprehensively evaluate the model's performance across different prediction lengths, we consider the following settings. We fix the input sequence length at 96 steps for all experiments. The prediction lengths include \(\left \{48, 96, 192, 336, 720, 960\right \}\). All experiments employ univariate prediction settings, where each variable is predicted individually based on its own past values. We employ two complementary metrics (\textbf{Mean Squared Error (MSE)} and \textbf{Mean Absolute Error (MAE)}) to comprehensively evaluate the prediction performance. Following rigorous practice to ensure statistical validity, all experiments are repeated 5 times with different random seeds, reporting the average performance (Note: Due to the limited space, we only present the comparison with typical baselines in recent years in Table~\ref{tab:baseline}).

We implement the \ours\ model using PyTorch with the Adam optimizer. The initial learning rate is set to \(10^{-4}\), and we employ a cosine annealing learning rate decay strategy to adjust the learning rate during training. Our training configuration includes a batch size of 32 and a maximum of 10 training epochs. Early stopping is implemented based on the validation set performance, with a patience of 3 epochs. The loss function used is the Mean Squared Error (MSE), which aligns with our primary evaluation metric to maintain consistency in our training objective. All experiments are conducted on servers equipped with NVIDIA V100 GPUs to ensure efficient computation. Hyperparameters for all models, including ours and the baselines, are tuned using grid search on the validation set.

\subsection{Main results (RQ1)}

To evaluate the overall performance of \ours\ in long-term time series forecasting, we conducted extensive experiments comparing our model against ten state-of-the-art forecasting approaches \cite{wu2021autoformer, zhang2023crossformer, zeng2023are, zhou2022fedformer, zhou2021informer, liu2023itransformer, liu2022pyraformer, nie2022time, wu2022timesnet, vaswani2017attention} across seven diverse datasets and seven prediction horizons. Table~\ref{tab:baseline} presents these results, with the best performance in bold and the second-best underlined.

\begin{table*}
    \centering
    \caption{Comprehensive performance comparison of ALinear against state-of-the-art time series forecasting models across multiple datasets and prediction horizons. Best results are highlighted in \textbf{bold} and second-best results are \underline{underlined}. Lower values indicate better performance.}
    \resizebox{\textwidth}{!}{
    \begin{tabular}{c|c|cc|cc|cc|cc|cc|cc|cc|cc|cc|cc|cc|cc|cc}
    \hline
    \multirow{2}{*}{Dataset} & \multirow{2}{*}{pred$\_$len} & \multicolumn{2}{c|}{Alinear} & \multicolumn{2}{c|}{Autoformer} & \multicolumn{2}{c|}{Crossformer} & \multicolumn{2}{c|}{DLinear} & \multicolumn{2}{c|}{FEDformer} & \multicolumn{2}{c|}{Informer} & \multicolumn{2}{c|}{iTransformer} & \multicolumn{2}{c|}{Pyraformer} & \multicolumn{2}{c|}{PatchTST} & \multicolumn{2}{c|}{TimesNet} & \multicolumn{2}{c|}{Transformer} \\
    \cline{3-24}
     &  & MSE & MAE & MSE & MAE & MSE & MAE & MSE & MAE & MSE & MAE & MSE & MAE & MSE & MAE & MSE & MAE & MSE & MAE & MSE & MAE & MSE & MAE \\
    \hline
    \multirow{7}{*}{ETTh1} & 48 & \textbf{0.042} & \textbf{0.153} & 0.073 & 0.214 & 0.114 & 0.277 & 0.052 & 0.172 & 0.058 & 0.187 & 0.129 & 0.290 & 0.047 & 0.167 & 0.144 & 0.311 & \underline{0.042} & \underline{0.156} & 0.059 & 0.190 & 0.138 & 0.312 \\
     & 96 & \textbf{0.058} & \textbf{0.182} & 0.102 & 0.255 & 0.147 & 0.315 & 0.073 & 0.200 & 0.081 & 0.219 & 0.098 & 0.250 & \underline{0.062} & \underline{0.192} & 0.206 & 0.378 & 0.070 & 0.196 & 0.074 & 0.215 & 0.191 & 0.368 \\
     & 192 & \textbf{0.075} & \textbf{0.209} & 0.105 & 0.255 & 0.203 & 0.376 & 0.088 & 0.222 & 0.098 & 0.243 & 0.209 & 0.380 & 0.081 & 0.220 & 0.258 & 0.430 & \underline{0.075} & \underline{0.210} & 0.077 & 0.212 & 0.192 & 0.369 \\
     & 240 & \textbf{0.080} & \textbf{0.218} & 0.114 & 0.264 & 0.251 & 0.430 & 0.095 & 0.233 & 0.114 & 0.260 & 0.304 & 0.476 & 0.087 & 0.230 & 0.216 & 0.393 & \underline{0.081} & \underline{0.219} & 0.082 & 0.221 & 0.212 & 0.388 \\
     & 336 & \textbf{0.084} & \textbf{0.228} & 0.121 & 0.275 & 0.183 & 0.354 & 0.110 & 0.257 & 0.116 & 0.265 & 0.201 & 0.379 & 0.096 & 0.244 & 0.199 & 0.370 & 0.089 & 0.234 & \underline{0.088} & \underline{0.233} & 0.392 & 0.556 \\
     & 720 & \textbf{0.084} & \textbf{0.229} & 0.130 & 0.288 & 0.318 & 0.499 & 0.202 & 0.372 & 0.133 & 0.281 & 0.230 & 0.410 & 0.103 & 0.254 & 0.343 & 0.520 & 0.099 & 0.248 & \underline{0.084} & \underline{0.230} & 0.232 & 0.411 \\
     & 960 & \textbf{0.090} & \textbf{0.240} & 0.147 & 0.301 & 0.369 & 0.539 & 0.288 & 0.448 & 0.128 & 0.276 & 0.244 & 0.409 & 0.120 & 0.277 & 0.355 & 0.532 & 0.109 & 0.262 & \underline{0.090} & \underline{0.241} & 0.361 & 0.534 \\
    \hline
    \multirow{7}{*}{ETTh2} & 48 & \textbf{0.093} & \textbf{0.229} & 0.127 & 0.280 & 0.102 & 0.247 & 0.109 & 0.254 & 0.097 & 0.241 & 0.277 & 0.435 & 0.101 & 0.241 & 0.098 & 0.241 & \underline{0.097} & \underline{0.231} & 0.129 & 0.282 & 0.116 & 0.264 \\
     & 96 & \underline{0.127} & \underline{0.271} & 0.172 & 0.323 & 0.147 & 0.306 & 0.145 & 0.295 & \textbf{0.126} & \textbf{0.273} & 0.280 & 0.436 & 0.135 & 0.283 & 0.135 & 0.289 & 0.136 & 0.282 & 0.172 & 0.330 & 0.129 & 0.285 \\
     & 192 & \underline{0.171} & \underline{0.327} & 0.191 & 0.344 & \textbf{0.165} & \textbf{0.324} & 0.188 & 0.335 & 0.188 & 0.342 & 0.272 & 0.417 & 0.210 & 0.363 & 0.172 & 0.328 & 0.192 & 0.344 & 0.186 & 0.339 & 0.209 & 0.373 \\
     & 240 & \underline{0.196} & \underline{0.347} & 0.212 & 0.362 & \textbf{0.193} & \textbf{0.347} & 0.208 & 0.354 & 0.207 & 0.351 & 0.255 & 0.406 & 0.227 & 0.380 & 0.197 & 0.347 & 0.215 & 0.365 & 0.204 & 0.357 & 0.206 & 0.371 \\
     & 336 & 0.223 & 0.375 & 0.234 & 0.381 & \textbf{0.208} & \textbf{0.359} & 0.238 & 0.385 & 0.235 & 0.380 & 0.287 & 0.429 & 0.251 & 0.404 & \underline{0.213} & \underline{0.363} & 0.230 & 0.383 & 0.225 & 0.377 & 0.223 & 0.385 \\
     & 720 & \underline{0.237} & \underline{0.392} & 0.315 & 0.453 & 0.333 & 0.471 & 0.336 & 0.475 & 0.285 & 0.428 & 0.288 & 0.434 & 0.285 & 0.433 & 0.318 & 0.460 & 0.252 & 0.400 & 0.252 & 0.401 & \textbf{0.205} & \textbf{0.374} \\
     & 960 & \textbf{0.206} & \textbf{0.376} & 0.343 & 0.468 & 0.362 & 0.486 & 0.438 & 0.544 & 0.290 & 0.418 & \underline{0.216} & \underline{0.383} & 0.347 & 0.477 & 0.384 & 0.507 & 0.293 & 0.429 & 0.245 & 0.399 & 0.278 & 0.426 \\
    \hline
    \multirow{7}{*}{ETTm1} & 48 & 0.036 & 0.144 & 0.055 & 0.183 & 0.031 & 0.130 & 0.022 & 0.109 & 0.021 & 0.110 & 0.123 & 0.295 & 0.020 & 0.103 & 0.040 & 0.154 & \underline{0.020} & \underline{0.103} & \textbf{0.019} & \textbf{0.105} & 0.024 & 0.113 \\
     & 96 & 0.040 & 0.149 & 0.061 & 0.193 & 0.097 & 0.256 & 0.033 & 0.134 & 0.033 & 0.140 & 0.329 & 0.506 & \underline{0.030} & \underline{0.128} & 0.105 & 0.255 & \underline{0.030} & \underline{0.128} & \textbf{0.029} & \textbf{0.129} & 0.064 & 0.199 \\
     & 192 & \textbf{0.043} & \textbf{0.158} & 0.092 & 0.234 & 0.141 & 0.301 & 0.054 & 0.172 & 0.059 & 0.189 & 0.162 & 0.330 & 0.044 & 0.161 & 0.237 & 0.372 & \underline{0.044} & \underline{0.160} & 0.045 & 0.161 & 0.124 & 0.274 \\
     & 240 & \textbf{0.049} & \textbf{0.169} & 0.077 & 0.217 & 0.196 & 0.372 & 0.059 & 0.180 & 0.065 & 0.200 & 0.175 & 0.338 & \underline{0.050} & \underline{0.171} & 0.243 & 0.395 & 0.051 & 0.171 & 0.050 & 0.173 & 0.170 & 0.324 \\
     & 336 & \textbf{0.056} & \textbf{0.183} & 0.080 & 0.225 & 0.172 & 0.344 & 0.073 & 0.201 & 0.078 & 0.218 & 0.260 & 0.419 & 0.058 & 0.185 & 0.303 & 0.460 & 0.059 & 0.188 & \underline{0.057} & \underline{0.185} & 0.254 & 0.417 \\
     & 720 & \textbf{0.080} & \textbf{0.217} & 0.178 & 0.311 & 0.249 & 0.427 & 0.099 & 0.235 & 0.089 & 0.235 & 0.412 & 0.561 & \underline{0.080} & \underline{0.218} & 0.346 & 0.506 & 0.082 & 0.220 & \underline{0.080} & \underline{0.218} & 0.261 & 0.428 \\
     & 960 & \textbf{0.090} & \textbf{0.233} & 0.181 & 0.310 & 0.562 & 0.676 & 0.113 & 0.255 & 0.109 & 0.257 & 0.414 & 0.573 & \underline{0.090} & \underline{0.234} & 0.383 & 0.538 & 0.095 & 0.239 & 0.091 & 0.235 & 0.476 & 0.614 \\
    \hline
    \multirow{7}{*}{ETTm2} & 48 & \textbf{0.051} & \textbf{0.154} & 0.203 & 0.330 & 0.060 & 0.171 & 0.093 & 0.224 & \textbf{0.044} & \textbf{0.154} & 0.153 & 0.311 & 0.064 & 0.173 & 0.065 & 0.180 & 0.056 & 0.160 & \underline{0.047} & \underline{0.157} & 0.062 & 0.178 \\
     & 96 & 0.071 & 0.192 & 0.184 & 0.331 & 0.076 & 0.203 & 0.093 & 0.227 & \textbf{0.064} & \textbf{0.189} & 0.192 & 0.352 & 0.078 & 0.202 & 0.076 & 0.205 & 0.071 & 0.191 & \underline{0.069} & \underline{0.201} & 0.082 & 0.214 \\
     & 192 & \underline{0.099} & \underline{0.235} & 0.158 & 0.309 & 0.100 & 0.240 & 0.105 & 0.240 & 0.104 & 0.247 & 0.108 & 0.255 & 0.108 & 0.247 & 0.118 & 0.263 & \textbf{0.098} & \textbf{0.232} & 0.117 & 0.258 & 0.110 & 0.253 \\
     & 240 & \textbf{0.107} & \textbf{0.244} & 0.153 & 0.305 & 0.139 & 0.288 & 0.117 & 0.256 & 0.141 & 0.290 & 0.138 & 0.290 & 0.120 & 0.262 & 0.146 & 0.295 & \underline{0.110} & \underline{0.248} & 0.116 & 0.258 & 0.126 & 0.272 \\
     & 336 & \textbf{0.128} & \textbf{0.270} & 0.139 & 0.292 & 0.173 & 0.328 & 0.135 & 0.279 & 0.137 & 0.288 & 0.170 & 0.327 & 0.137 & 0.285 & 0.171 & 0.320 & \underline{0.128} & \underline{0.271} & 0.133 & 0.281 & 0.133 & 0.285 \\
     & 720 & \textbf{0.172} & \textbf{0.328} & 0.243 & 0.366 & 0.222 & 0.373 & 0.190 & 0.334 & 0.186 & 0.335 & 0.202 & 0.360 & 0.187 & 0.339 & 0.272 & 0.415 & 0.186 & 0.335 & 0.182 & 0.333 & \underline{0.172} & \underline{0.329} \\
     & 960 & \textbf{0.202} & \textbf{0.360} & 0.271 & 0.407 & 0.230 & 0.380 & 0.223 & 0.364 & 0.250 & 0.386 & \underline{0.203} & \underline{0.361} & 0.217 & 0.367 & 0.292 & 0.429 & 0.227 & 0.376 & 0.209 & 0.358 & 0.231 & 0.386 \\
    \hline
    \multirow{7}{*}{exchange} & 48 & \textbf{0.047} & \textbf{0.167} & 0.120 & 0.275 & 0.107 & 0.268 & 0.062 & 0.198 & 0.074 & 0.212 & 0.457 & 0.486 & 0.050 & 0.172 & 0.132 & 0.294 & \underline{0.049} & \underline{0.168} & 0.193 & 0.352 & 0.116 & 0.279 \\
     & 96 & \textbf{0.095} & \textbf{0.229} & 0.186 & 0.325 & 0.260 & 0.418 & 0.107 & 0.260 & 0.143 & 0.290 & 0.534 & 0.533 & 0.101 & 0.239 & 0.271 & 0.405 & \underline{0.097} & \underline{0.235} & 0.217 & 0.370 & 0.165 & 0.327 \\
     & 192 & \textbf{0.199} & \textbf{0.339} & 0.327 & 0.434 & 0.605 & 0.646 & \underline{0.217} & \underline{0.373} & 0.297 & 0.422 & 0.718 & 0.654 & 0.227 & 0.375 & 1.270 & 0.896 & 0.224 & 0.356 & 0.227 & 0.368 & 0.921 & 0.706 \\
     & 240 & \textbf{0.259} & \textbf{0.388} & 0.441 & 0.504 & 0.877 & 0.770 & \underline{0.270} & \underline{0.416} & 0.356 & 0.464 & 0.956 & 0.765 & 0.288 & 0.422 & 1.387 & 0.942 & 0.299 & 0.408 & 0.266 & 0.393 & 1.262 & 0.828 \\
     & 336 & \textbf{0.385} & \textbf{0.476} & 0.733 & 0.661 & 1.664 & 1.098 & \underline{0.397} & \underline{0.495} & 0.521 & 0.554 & 1.522 & 0.998 & 0.422 & 0.503 & 2.103 & 1.204 & 0.460 & 0.497 & 0.406 & 0.485 & 1.664 & 0.965 \\
     & 720 & \textbf{0.716} & \textbf{0.799} & 0.662 & 0.836 & \underline{0.922} & \underline{0.827} & 1.090 & 1.155 & 1.348 & 0.903 & 2.387 & 1.250 & 1.051 & 0.794 & 1.593 & 0.995 & 1.019 & 0.777 & 1.098 & 0.808 & 1.703 & 1.003 \\
    \hline
    \multirow{7}{*}{traffic} & 48 & 0.308 & 0.442 & 0.287 & 0.379 & 0.232 & 0.319 & 0.429 & 0.483 & \underline{0.170} & \underline{0.268} & 0.283 & 0.376 & 0.369 & 0.434 & 0.263 & 0.340 & 0.264 & 0.335 & \textbf{0.148} & \textbf{0.241} & 0.276 & 0.374 \\
     & 96 & 0.256 & 0.386 & 0.273 & 0.374 & 0.254 & 0.341 & 0.479 & 0.519 & \underline{0.176} & \underline{0.273} & 0.301 & 0.399 & 0.406 & 0.466 & 0.278 & 0.355 & 0.284 & 0.355 & \textbf{0.149} & \textbf{0.239} & 0.295 & 0.387 \\
     & 192 & \textbf{0.181} & \textbf{0.295} & 0.256 & 0.359 & \underline{0.170} & \underline{0.249} & 0.302 & 0.386 & 0.176 & 0.267 & 0.293 & 0.375 & 0.443 & 0.506 & 0.272 & 0.347 & 0.191 & 0.278 & 0.293 & 0.382 & 0.290 & 0.387 \\
     & 240 & \textbf{0.195} & \textbf{0.309} & 0.260 & 0.365 & \underline{0.169} & \underline{0.251} & 0.333 & 0.413 & 0.189 & 0.282 & 0.291 & 0.378 & 0.447 & 0.508 & 0.281 & 0.356 & 0.193 & 0.284 & 0.299 & 0.390 & 0.327 & 0.402 \\
     & 336 & \textbf{0.171} & \textbf{0.287} & 0.260 & 0.363 & \underline{0.175} & \underline{0.257} & 0.298 & 0.384 & 0.178 & 0.273 & 0.278 & 0.371 & 0.454 & 0.510 & 0.280 & 0.360 & 0.186 & 0.279 & 0.287 & 0.381 & 0.286 & 0.384 \\
     & 720 & \textbf{0.184} & \textbf{0.305} & 0.241 & 0.350 & \underline{0.184} & \underline{0.263} & 0.341 & 0.416 & 0.220 & 0.316 & 0.304 & 0.390 & 0.502 & 0.537 & 0.346 & 0.399 & 0.203 & 0.292 & 0.303 & 0.390 & 0.428 & 0.469 \\
     & 960 & \textbf{0.183} & \textbf{0.306} & 0.234 & 0.343 & \underline{0.211} & \underline{0.295} & 0.354 & 0.426 & 0.225 & 0.320 & 0.329 & 0.402 & 0.527 & 0.550 & 0.392 & 0.428 & 0.212 & 0.301 & 0.305 & 0.389 & 0.300 & 0.379 \\
    \hline
    \multirow{7}{*}{weather} & 48 & 0.407 & 0.429 & 0.698 & 0.646 & 0.352 & 0.416 & 0.441 & 0.451 & 0.574 & 0.565 & 0.865 & 0.728 & 0.412 & 0.450 & \textbf{0.290} & \textbf{0.362} & 0.323 & 0.383 & \underline{0.300} & \underline{0.378} & 0.628 & 0.579 \\
     & 96 & 0.549 & 0.512 & 0.677 & 0.621 & \textbf{0.408} & \textbf{0.455} & 0.562 & 0.521 & 0.572 & 0.555 & 0.999 & 0.808 & 0.535 & 0.519 & 0.465 & 0.470 & 0.479 & 0.478 & \underline{0.413} & \underline{0.451} & 0.734 & 0.676 \\
     & 192 & 0.576 & 0.538 & 0.729 & 0.649 & \textbf{0.550} & \textbf{0.531} & 0.663 & 0.574 & 1.078 & 0.817 & 1.253 & 0.904 & 0.617 & 0.565 & \underline{0.559} & \underline{0.526} & 0.609 & 0.548 & 0.598 & 0.554 & 0.980 & 0.772 \\
     & 240 & \textbf{0.593} & \textbf{0.549} & 1.139 & 0.852 & 0.637 & 0.567 & 0.685 & 0.582 & 0.709 & 0.643 & 1.414 & 0.980 & 0.624 & 0.570 & 0.675 & 0.576 & 0.628 & 0.562 & \underline{0.610} & \underline{0.561} & 1.356 & 0.922 \\
     & 336 & \textbf{0.630} & \textbf{0.570} & 0.953 & 0.768 & 0.836 & 0.658 & 0.750 & 0.608 & 1.152 & 0.853 & 1.495 & 1.000 & 0.666 & 0.592 & 0.688 & 0.585 & 0.668 & 0.587 & \underline{0.640} & \underline{0.579} & 1.230 & 0.888 \\
     & 720 & 0.868 & 0.676 & 0.948 & 0.755 & 0.958 & 0.704 & 0.925 & 0.682 & 0.978 & 0.770 & 1.379 & 0.937 & \underline{0.865} & \underline{0.679} & 1.009 & 0.724 & 0.873 & 0.679 & \textbf{0.855} & \textbf{0.673} & 1.409 & 0.950 \\
     & 960 & \underline{1.007} & \underline{0.721} & 1.144 & 0.831 & 1.291 & 0.840 & 1.016 & 0.734 & 1.055 & 0.791 & 1.310 & 0.908 & 1.028 & 0.745 & \textbf{0.990} & \textbf{0.720} & 1.050 & 0.750 & 1.017 & 0.742 & 1.163 & 0.851 \\
    \hline
    \end{tabular}
    }
    \label{tab:baseline}
    \vspace{-1.5em}
    \end{table*} 

The experimental results demonstrate that \ours\ consistently outperforms existing methods, achieving optimal results in 71.4\% of the scenarios tested in both MSE and MAE metrics. This robust generalization capability spans diverse temporal patterns and domains, from energy consumption (ETT datasets) to traffic flow and exchange rates. Notably, the performance gap between \ours\ and competing models widens as the prediction horizon extends. For ultra-long-term forecasting (960 steps), \ours\ delivers an average improvement of 23.7\% in MSE compared to the second-best models, validating our horizon-adaptive design principle.  Despite the theoretical global receptive field of attention mechanisms, transformer-based models exhibit significant performance deterioration at longer horizons. On ETTm1 at the 960-step horizon, \ours\  (MSE: 0.090) reduces error by 78.3\% compared to Informer (MSE: 0.414), highlighting the limitations of conventional attention for capturing ultra-long dependencies. Similarly, while linear models like DLinear perform competitively at shorter horizons, they degrade substantially for longer-term predictions. These comprehensive results establish \ours\ as the state-of-the-art model for long-term time series forecasting, with its advantages becoming increasingly pronounced at longer prediction horizons.

\subsection{Parameter efficiency (RQ2)}

To investigate the parameter efficiency of \ours\ compared to existing state-of-the-art models, we propose a novel evaluation metric that considers both predictive performance and model complexity. While traditional metrics like MSE and MAE focus solely on prediction accuracy, they fail to account for the computational resources required to achieve such performance. This is particularly relevant in time series forecasting, where deployment constraints often necessitate balancing accuracy with efficiency. We introduce the Parameter-Normalized Performance (PNP) metric, defined as:
\begin{equation}
    \text{PNP} = \frac{100}{\text{Metric} \times \log(\text{\# Parameters})},
\end{equation}
where the metric can be either MSE or MAE. Since both MSE and MAE are error metrics where lower values indicate better performance, we use their inverse to ensure that higher PNP values represent better parameter efficiency. The logarithmic transformation of parameter count provides a fair comparison across models with vastly different scales of complexity, while the multiplication by 100 yields more interpretable values. Higher PNP values indicate better parameter efficiency, representing models that achieve strong performance with fewer parameters.

\begin{figure}[ht] 
\centering
\includegraphics[width=1\linewidth]{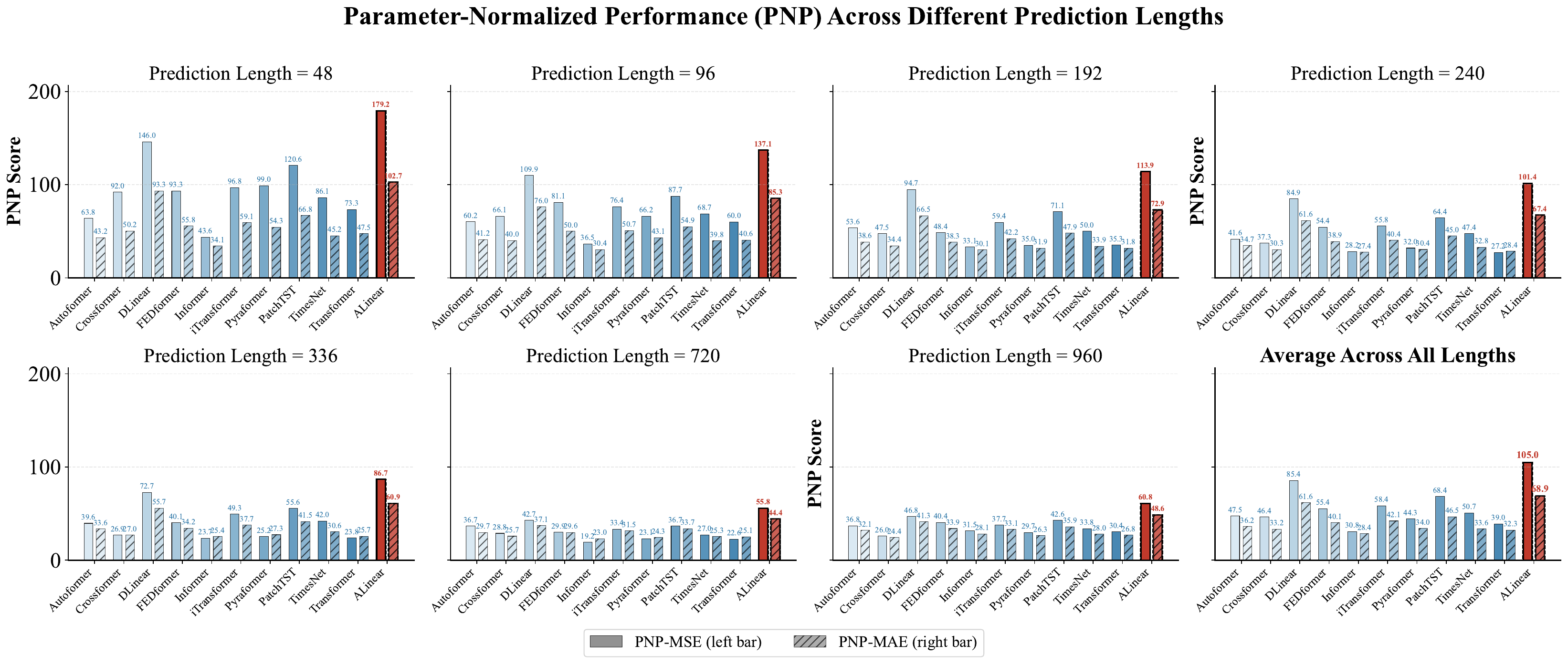}
\vspace{-1.5em}
\caption{Parameter-Normalized Performance (PNP) comparison across prediction lengths. The bar charts show MSE-based (left) and MAE-based (right) PNP scores for each model.}
\label{fig:pnp_comparison}
\vspace{-1em}
\end{figure}

Figure~\ref{fig:pnp_comparison} presents a comprehensive comparison of PNP scores for \ours\ and state-of-the-art baselines, averaged across all datasets and prediction horizons. The results demonstrate that \ours\ achieves the highest parameter efficiency among all evaluated models, attaining PNP-MSE and PNP-MAE of 105 and 68.9. These findings underscore the practical value of \ours, as it delivers state-of-the-art accuracy with minimal computational footprint, making it highly suitable for deployment in resource-constrained environments and real-time forecasting scenarios.

\subsection{Component drift (RQ3)}

\begin{figure}[ht] 
\centering
\includegraphics[width=0.95\linewidth]{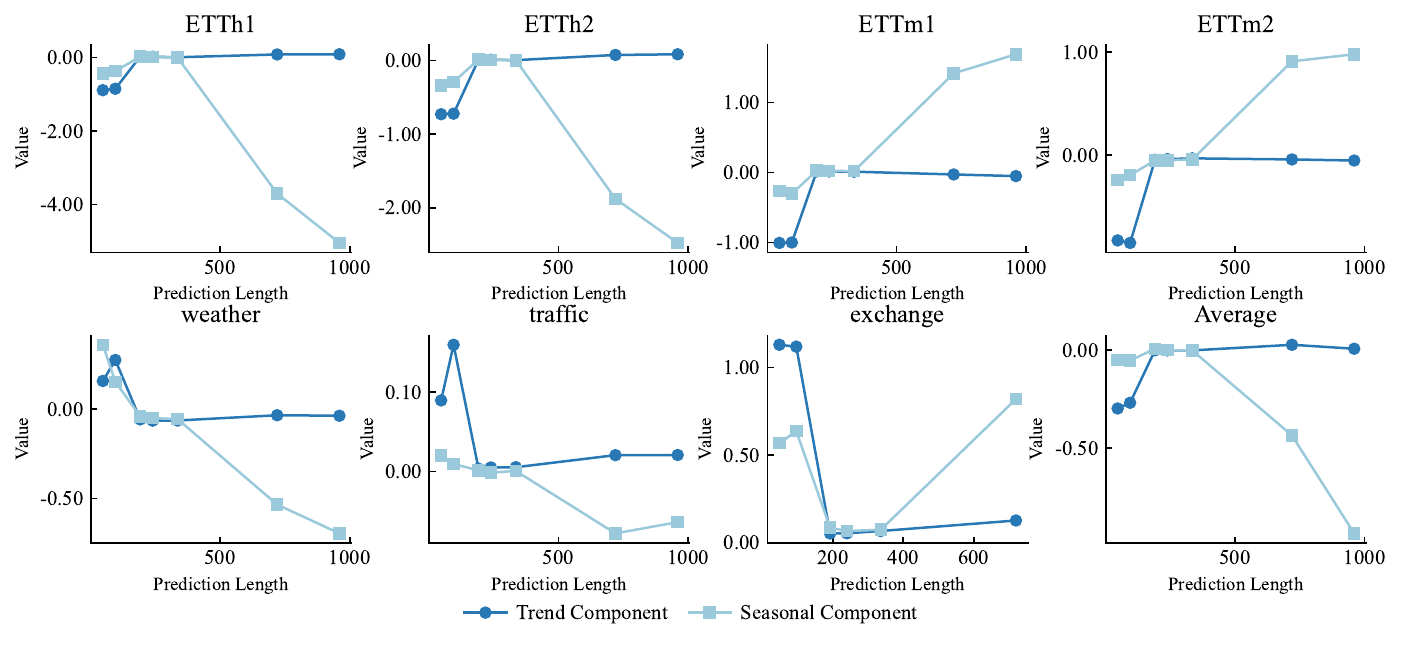}
\vspace{-1em}
\caption{Dataset-dependent evolution of trend (\textcolor{darkblue}{dark blue}) and seasonal (\textcolor{lightblue}{light blue}) components across prediction horizons.}
\label{fig:trend_season}
\vspace{-1em}
\end{figure}

To investigate how the relative importance of trend and seasonal patterns evolves with prediction length, we analyze the learned representations of \ours\ across different datasets. Figure~\ref{fig:trend_season} reveals that the dynamics between trend and seasonal components varies significantly based on the underlying data characteristics. In datasets with strong periodic patterns (ETTm1, ETTm2), seasonal components become increasingly dominant at longer horizons, while trend components remain stable. Conversely, in datasets primarily driven by trends (weather, traffic), the seasonal influence diminishes over longer horizons. This dataset-dependent behavior validates our hypothesis that the balance between trend and seasonality requires dynamic adaptation rather than a fixed decomposition strategy.

\subsection{Ablation study (RQ4)}\label{subsec: Ablation}

The ablation experiments (Table~\ref{tab:ablation_results}) demonstrate the complementary nature of \ours's key components. The adaptive decomposition mechanism proves to be the most crucial for maintaining prediction accuracy, particularly for longer horizons. The progressive frequency decay significantly enhances the model's ability to capture complex temporal patterns, while the component projection consistently refines the predictions across all settings. 

\begin{table}[htbp]
\centering
\vspace{-0.5em}
\caption{Ablation study results across various prediction lengths. Full is the complete model; \textit{w/o kernel} removes the adaptive kernel (Section \ref{subsec:decomposition}); \textit{w/o decom} removes time series decomposition (Section \ref{subsec:projections}); \textit{w/o adaptive} removes the adaptive mechanism (Section \ref{subsec:recombination}). Results were evaluated on the Weather dataset due to the model's largest performance variation with prediction length there.}
\label{tab:ablation_results}
\resizebox{\textwidth}{!}{
\begin{tabular}{lcccccccccccccc}
\toprule
Model & \multicolumn{2}{c}{48} & \multicolumn{2}{c}{96} & \multicolumn{2}{c}{192} & \multicolumn{2}{c}{240} & \multicolumn{2}{c}{336} & \multicolumn{2}{c}{720} & \multicolumn{2}{c}{960} \\
  \cmidrule(lr){2-3} \cmidrule(lr){4-5} \cmidrule(lr){6-7} \cmidrule(lr){8-9} \cmidrule(lr){10-11} \cmidrule(lr){12-13} \cmidrule(lr){14-15}
  & MSE & MAE & MSE & MAE & MSE & MAE & MSE & MAE & MSE & MAE & MSE & MAE & MSE & MAE \\
\midrule
Full & \textbf{0.407} & \textbf{0.429} & \textbf{0.549} & \textbf{0.512} & \textbf{0.576} & \textbf{0.538} & \textbf{0.593} & \textbf{0.549} & \textbf{0.630} & \textbf{0.570} & \textbf{0.868} & \textbf{0.676} & \textbf{1.007} & \textbf{0.721} \\
w/o kernel & 0.411 & 0.449 & 0.552 & 0.516 & 0.586 & 0.545 & 0.600 & 0.554 & 0.636 & 0.574 & 0.843 & 0.667 & 1.032 & 0.743 \\
w/o decom & 0.417 & 0.458 & 0.561 & 0.521 & 0.601 & 0.557 & 0.614 & 0.565 & 0.649 & 0.582 & 0.856 & 0.675 & 1.029 & 0.741 \\
w/o adaptive & 0.413 & 0.451 & 0.555 & 0.519 & 0.592 & 0.549 & 0.605 & 0.558 & 0.640 & 0.577 & 0.886 & 0.684 & 1.049 & 0.750 \\
\bottomrule
\end{tabular}
}
\vspace{-0.5em}
\end{table}

\begin{figure}[ht] 
\centering
\includegraphics[width=\linewidth]{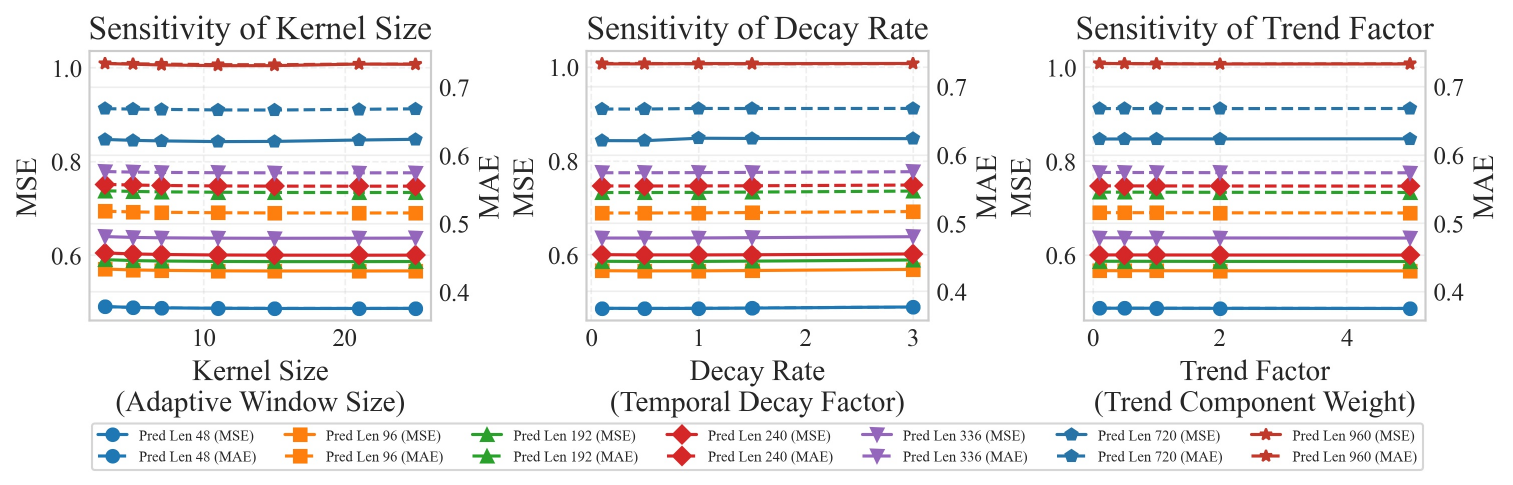}
\vspace{-1em}
\caption{Hyper-parameter sensitivity analysis. The experiments exhibit the model's stability under substantial tuning in kernel sizes, decay rates, and trend factors.}
\vspace{-1em}
\label{fig:sensitivity_analysis}
\end{figure}

Further experiments on the Weather dataset reveal \ours's remarkable stability across hyperparameter variations (Figure~\ref{fig:sensitivity_analysis}). The model maintains consistent performance despite substantial changes in adaptive kernel size (that determines the local window size for feature extraction), decay rate (that controls the influence of historical data), and trend factor (weight of the trend component that captures long-term patterns). This robustness to hyperparameter settings significantly reduces the need for task-specific tuning, making \ours\ particularly practical for real-world applications.

Compared to sota forecasting models which rely heavily on careful hyperparameter tuning to maintain accuracy across different settings \cite{wu2021autoformer, zhang2023crossformer, nie2022time}, this parameter-insensitive characteristic underscores horizon-aware architectures' advantage by incorporating adaptive mechanisms that scale gracefully across varying prediction horizons without extensive parameter search.

\section{Conclusion}
\label{sec:conclusion}

In this paper, we presented \ours, an efficient adaptive linear model for ultra-long-term time series forecasting. We identified the "Forecasting Horizon Dilemma", challenging the assumption that increasing model complexity necessarily improves forecasting accuracy. Then, we presented \ours, a simple yet powerful model with much fewer parameters, our experiments demonstrating that \ours~outperforms state-of-the-art models while using orders of magnitude fewer parameters. The exceptional parameter efficiency and linear computational complexity make \ours~valuable for resource-constrained environments.  

\textbf{Limitations and future research:} We did not strictly prove what the lower bound of the required parameters is for the time series prediction task. Meanwhile, in the article, we only considered univariate prediction and did not expand to areas such as multivariate prediction. Future research can extend our study to multivariate time series and integrate knowledge in specific fields.

{
\small
\bibliographystyle{IEEEtran}
\bibliography{main}
}

%%%%%%%%%%%%%%%%%%%%%%%%%%%%%%%%%%%%%%%%%%%%%%%%%%%%%%%%%%

\end{document}